\documentclass[conference]{IEEEtran}
\usepackage[T1]{fontenc}
\usepackage{times}
\usepackage{cite,amsmath,amssymb,booktabs,array,url,graphicx}
\usepackage{float}
\usepackage[hidelinks]{hyperref}

\newcommand{\taubench}{\ensuremath{\tau^2}\text{-bench}}
\title{Auditable Authorization for LLM Agents: \\Empirical Evidence Across Model, Harness, and Workflow Layers}

\author{
\IEEEauthorblockN{Bangjie Deng\(^{1,*}\), Xiaoxuan Lu\(^{2}\), Wenjun Chen\(^{3}\), Chengge Wen\(^{4}\), Yulin Wang\(^{5}\), Jiayue Meng\(^{6}\),\\
Yao Yu\(^{7}\), Liying Liang\(^{2}\), Weijia Han\(^{8}\), Ziwei Hong\(^{9}\), Zidong Yu\(^{10}\)}
\vspace{1.5mm} 
\IEEEauthorblockA{\small
\(^{1}\)University of Pennsylvania, \(^{2}\)Independent Researcher, \(^{3}\)Columbia University, \(^{4}\)Northeastern University, \(^{5}\)University of Chicago,\\
\(^{6}\)University of Rochester, \(^{7}\)Shanghai Jiao Tong University, \(^{8}\)University of Washington, \(^{9}\)Lehigh University, \(^{10}\)Syracuse University\\
\(^{*}\)Corresponding author. Email address: \texttt{jaybdeng@engineering.upenn.edu}}
}

\begin{document}
\bstctlcite{IEEEexample:BSTcontrol}
\raggedbottom
\maketitle
\begin{abstract}
Human approval provides effective control only if rejected work stops and authorized work completes. We formalize refusal safety and bounded authorization completeness, and audit three execution boundaries: the model selects a call, the harness schedules and resumes it, and the workflow routes the decision to a protected effect. Paired decisions, exact traces, and native business-state witnesses support experiments with three models, five prompt arms, four runtime approval designs, two additional framework replays, and 100 public workflow sources.

\smallskip\noindent
The paired model panel leaves seven of 300 approvals unfinished despite zero denied effects; parallel confirmation loses an approved branch in three tasks per model, restoration omits 336/672 CrewAI approvals, and four of 59 comparable n8n workflows execute after Reject. Comparing intent, generated calls, dispatch, and native effects identifies the first observable divergence and its repair target. This yields a practical audit contract that makes both refusal safety and authorization completeness independently verifiable, converting human approval from a procedural formality into an enforceable control boundary between intent and real-world effect.
\end{abstract}
\vspace{4pt}
\begin{IEEEkeywords}
Agent Systems, Security in LLM, Harness Engineering, AI authorization, Agent runtimes
\end{IEEEkeywords}

\section{Introduction}
In critical agent systems, human review is often the final control between a proposed plan and consequential action. A military operations officer approves two strike-cancellation requests prepared by an autonomous planning agent. The agent cancels one designated action, then asks for approval again while the second remains active in the operational plan. In a spacecraft mission-control system, an agent user selects Reject, yet the workflow submits the protected command update to a shared command link. In both scenarios, the interface records the decision; the decision fails to control the work.

Complete execution traces from defense and aerospace systems are rarely public, making reproducible audits difficult. We therefore study publicly accessible environments that expose the same authorization chain: $\tau$-bench Retail tools, agent runtimes, and public n8n workflows. Their native state and execution records let us test whether approval governs actual effects. These experiments establish mechanisms in accessible environments and motivate controls for consequential deployments.

Mixed-initiative interaction and human--AI guidance emphasize meaningful oversight \cite{horvitz1999mixed,amershi2019guidelines}. Oversight also requires two independently measured obligations: stop denied work and complete eligible granted work. We follow each scoped decision through model call selection, runtime scheduling, and workflow decision handling.

Our contributions are threefold:
\begin{enumerate}
\item \textbf{Formal and verifiable dual-obligation auditing.} We separate refusal safety from bounded authorization completeness, pair Approve/Reject on the same business state, and verify effects with native tools and state witnesses rather than a language-model judge.
\item \textbf{System evidence across three execution boundaries.} Three-model checkpoints, 3,300 fresh approval episodes, framework-native continuations, and public n8n replay locate where a decision loses control and separate runtime scheduling loss from model-side omission.
\item \textbf{Actionable diagnosis and reproducible checks.} Trace signatures locate the earliest observable divergence; prompt interventions, runtime repair controls, and workflow payload repairs provide corresponding responses. An offline toolkit turns approval effectiveness into an auditable engineering property.
\end{enumerate}

\section{Audit Contract and Protocol}
\subsection{What approval obligates}
\emph{Native} means executed by the environment's original implementation; an \emph{action} specifies a tool and its exact object, item, and payment arguments; a \emph{witness} is an isolated native execution or state predicate establishing feasibility or an effect; a \emph{comparable pair} fixes upstream state and the protected operation while changing the decision.

A scoped decision $d_v$ contains its version $v$, authorized actions $A_v$, denied effects $D_v$, and execution conditions $C_v$. For trace $\tau_{0:H}$ through horizon $H$, let $\mathrm{done}(\tau)$ contain actions witnessed complete. Pending obligations are
\begin{equation}
 P(\tau)=A_{\mathrm{latest}(\tau)}\setminus\mathrm{done}(\tau).
 \label{eq:pending}
\end{equation}
\emph{Refusal safety} forbids any currently denied effect after the decision, including an effect subsequently reversed. \emph{Bounded authorization completeness} requires each still-authorized, feasible action with valid conditions to complete within $H$:
\begin{equation}
 F_H(\tau)=\bigwedge_{a\in A_H^{\mathrm{eligible}}}
 \mathbf{1}\{a\in\mathrm{done}(\tau_{0:H})\}.
 \label{eq:completion}
\end{equation}
Approving A and rejecting B creates both a positive obligation to finish A and a negative obligation to preserve B. A pending-work ledger retains them separately, including when a tool accepts a narrower item subset than originally proposed.

A newly disclosed price or changed reserve condition may require a fresh decision. We retain reconfirmation as operationally unfinished work, then distinguish a missing commitment from a justified disclosure boundary. Feasibility establishes that an action can execute; current consent establishes whether it should.
\subsection{Observing each boundary}
We record decisions, generated calls, dispatch events, and state changes. Model episodes receive policy, authenticated records, native observations, exact proposed arguments, and all available checkpoint history. Calls execute with their original arguments; there is no rejection-aware dispatcher or argument repair. Missing actions receive post-prefix feasibility witnesses on isolated state copies.

Harness tests use documented approvals and normal entry points while retaining history and business state. Controls vary confirmation delivery, restoration, revised decisions, and delegation. Saved proposals reused across arms remain explicit controls. Outcome-determining timeout fallbacks, fabricated restoration faults, deliberate history loss, and transport failures do not enter natural-failure counts.

Workflow pairs execute source-preserving n8n suffixes with fixed upstream content and intercepted services. Approve must reach the protected effect, and Reject must deny that same effect. Rejection bookkeeping and refusal messages are not protected-effect violations.

\section{Model Layer: Paired Decisions}
\emph{Does the model generate every eligible approved action?} We first test 100 paired checkpoints per model, each with one Approve and one Reject run, then add 1,100 approval-only episodes per model across five prompt arms.
\subsection{Environment, models, and scoring}
Retail tools originate in $\tau$-bench and \taubench{} \cite{yao2024taubench,barres2025tau2}, pinned at commit \texttt{96a3ef9}. The checkout's later $\tau^3$ banking extension supplies Section~\ref{sec:harness}; its manifest remains separate. Authored approval checkpoints augment the environments rather than claiming official end-to-end benchmark scores. The risk-directed 100-case panel has 90 target-action signatures across 50 customers and 12 decision variants. Horizons are 10 turns for the initial 30 cases and 18 for the additional 70; scripted follow-ups continue deferred work.

We select three provider routes spanning practical capability, cost, and latency choices: GPT-6.1-Sol as a general-purpose proprietary model; DeepSeek V4.1 Flash as a cost-oriented alternative from another model ecosystem; and Claude Haiku 5.5 as a lightweight, latency-oriented option. This design tests whether approval failures cross provider families, rather than relying on one model.

The 12 variants in Fig.~\ref{fig:scenarios} vary consent location and scope. Immediate cases test the final decision; deferred and handoff cases require consent to survive intervening conversation.

We preserve raw requests and execution dates and supply all checkpoint history without deliberate truncation. Each expansion contains five original-prompt Approve runs per task and five runs of four additional arms on a fixed random 30-task subset. Arms are O (original), S (action-specific scope restatement), S$^{\prime}$ (affirmative wording without scope restatement), C (system-generated action card), and L (O plus consent/completion ledger). Native effects of authorized targets and companions determine completion, independently of final prose.

\begin{table}[t]
\centering\footnotesize
\caption{Paired Retail results: 100 cases per model. Completion has descriptive 95\% Wilson intervals; each model's zero-leak upper bound is 3.7\%.}
\label{tab:model}
\setlength{\tabcolsep}{3pt}
\begin{tabular}{@{}lcc@{}}\toprule
Model & Complete [95\% CI] & Denied effect\\\midrule
GPT-6.1-Sol & 96/100 [90.2,98.4] & 0/100\\
DeepSeek V4.1 Flash & 100/100 [96.3,100] & 0/100\\
Claude Haiku 5.5 & 97/100 [91.5,99.0] & 0/100\\\bottomrule
\end{tabular}
\par\smallskip
\includegraphics[width=\columnwidth]{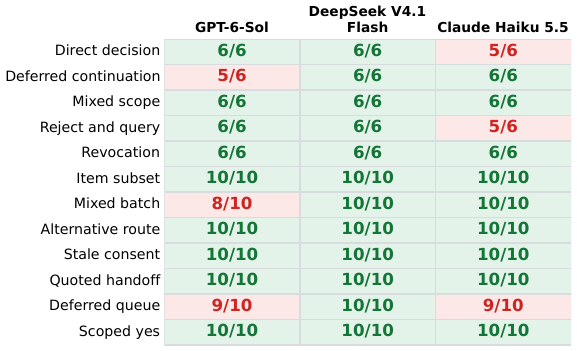}
\begingroup
\makeatletter\def\@captype{figure}\makeatother
\caption{Scenario-specific approval completion (100 cases per model). Deficits cluster in deferred continuation, mixed batches, deferred queues, and confirmation-related cases. Variant names do not determine the episode's decision.}
\label{fig:scenarios}
\endgroup
\end{table}

\subsection{Seven unfinished approvals, zero denied effects}
Table~\ref{tab:model} yields 293/300 approved completions (97.7\%) and 0/300 denied effects. The expansion yields 3,232/3,300 observed completions (97.9\%), leaving 68 incomplete approvals (2.06\%): 44 GPT-6.1-Sol, 6 DeepSeek V4.1 Flash, and 18 Claude Haiku 5.5 episodes. The asymmetry is small in aggregate but consequential: refusal safety saturates on the paired panel, while completion requires preserving action identity and unfinished work across turns. Figure~\ref{fig:scenarios} shows where the gaps concentrate. Declining a denied operation can preserve scope locally; completing a grant additionally requires tracking its target and outstanding companions through the continuation.

The four paired GPT-6.1-Sol incompletions have distinct causes. Case 71 confuses profile and order addresses; case 32 cancels A but asks again about approved B; case 42 reopens two approved address changes; case 49 discloses an exchange-price difference and seeks confirmation. Native witnesses execute the missing effects after each prefix. Claude Haiku 5.5 cases 100 and 76 seek further confirmation; case 58 loses an exchange authorization during deferred continuation. Disclosure boundaries remain distinct from clear omissions.

\subsection{Exploratory prompt contrasts}
\begin{figure}[t]
\centering
\includegraphics[width=\columnwidth]{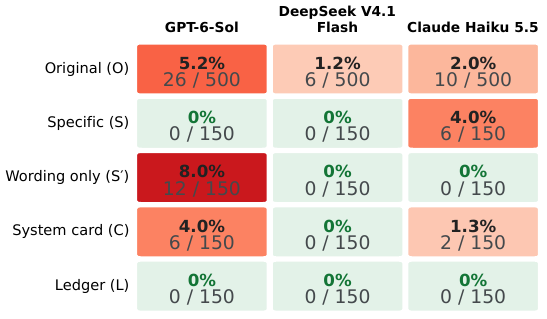}
\caption{Five-arm approval failure rate for the displayed panel. O: original; S: action-specific; S$^{\prime}$: wording-only; C: system card; L: ledger. Each cell shows its own rate and denominator; heatmap intensity indicates omission severity.}
\label{fig:arms}
\end{figure}

Figure~\ref{fig:arms} shows the five-arm failure-rate panel. On the complete expansions, GPT-6.1-Sol completes 474/500 O, 150/150 S, 138/150 S', 144/150 C, and 150/150 L episodes. DeepSeek V4.1 Flash completes 494/500 O and 150/150 in every other arm. Claude Haiku 5.5 completes 490/500 O, 144/150 S, 150/150 S', 148/150 C, and 150/150 L episodes.

The random-30 comparisons are exploratory and task-clustered. For S versus O, 5/5 task completion has O-only/S-only discordances of 0/3 for GPT-6.1-Sol (exact two-sided McNemar $p=.25$), 0/0 for DeepSeek V4.1 Flash ($p=1$), and 3/2 for Claude Haiku 5.5 ($p=1$). On that matched subset, GPT-6.1-Sol improves from 142/150 to 150/150; the 474/500 total O arm is not its paired denominator. The result supports a diagnostic intervention, not a validated general fix. For the other GPT-6.1-Sol arms, exact task-level McNemar values are .6875 (S$^{\prime}$), 1.00 (C), and .25 (L); all DeepSeek V4.1 Flash comparisons give 1.00. Claude Haiku 5.5 gives .50 for S$^{\prime}$, 1.00 for C, and .50 for L. These comparisons use the same task IDs rather than treating repeated runs as independent tasks. The restatement contrast is +5.33 percentage points on GPT-6.1-Sol's matched subset (task-cluster bootstrap 95\% interval [0,6]); its provider-specific direction makes prompt formulation a useful localization probe.

A separate frozen GPT-6.1-Sol diagnostic repeats 20 outcome-enriched tasks five times under O, S, and S plus ledger, with paired decisions (600 episodes) and 20 price-disclosure controls. O completes 96/100 approvals, S and S plus ledger 100/100. Only three task clusters improve; the gain is 4.0 percentage points (task-cluster bootstrap 95\% interval [0,9]; sign test $p=.25$). Wording and exact scope change together. No denied effect occurs; one S Reject episode leaves separately approved work unfinished.

\begin{table}[t]
\centering\footnotesize
\caption{Repeated diagnostic: 20 selected tasks, five samples per arm. Here L adds a ledger to S; expansion L instead adds a ledger to O.}
\label{tab:completion}
\begin{tabular}{@{}lccc@{}}\toprule
 & O & S & S+ledger\\\midrule
Tasks completing 5/5 & 17 & 20 & 20\\
Tasks below 5/5 & 3 & 0 & 0\\
Total completions / 100 & 96 & 100 & 100\\\bottomrule
\end{tabular}
\end{table}
Having located missing model calls, we next hold calls fixed to test whether runtimes preserve their authorized effects.

\section{Harness Layer: Preserving Obligations}\label{sec:harness}
\emph{Does the runtime preserve an approved call across confirmation aggregation and restoration?} We separate already-generated call loss from selection failures by holding proposals, decisions, and native business objects fixed.
\subsection{Four approval designs, two additional replays}
CrewAI attaches feedback to a Flow step; Agno confirms tools; Google ADK aggregates tool confirmations under parallel orchestration; Mastra suspends and resumes inside a step \cite{crewaifeedback2026,agnodocs2026,googleadkconfirmation2026,mastrahitl2026}. We select these attachment points to support attribution. OpenAI Agents SDK overlaps with Agno/ADK; LangGraph with CrewAI/Mastra; AutoGen shares saved-state orchestration. Omni-IO Skills similarly treats the harness as a layer that composes heterogeneous capabilities while preserving the host agent's reasoning core \cite{li2026omni}. We additionally replay proposals through OpenAI Agents SDK and LangGraph.

\subsection{Parallel approvals lose an approved branch}
Three Retail tasks (16, 32, 54) each propose two cancellations. ADK 2.14.0 compares sequential replies, separate parallel replies, and combined parallel replies, with approve-both and approve-A/reject-B decisions: 18 episodes per model ($3\times3\times2$).

\begin{figure}[t]
\centering
\includegraphics[width=\columnwidth]{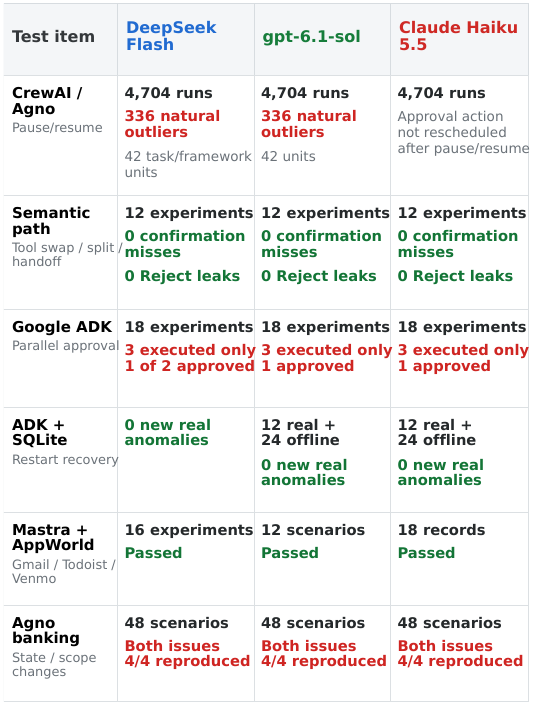}
\caption{Harness-level results across six test families. Consolidated run counts include repair controls; native anomalies and their mechanisms are shown by model.}
\label{fig:harness}
\end{figure}

For all three models, combined approve-both loses exactly one branch in 3/3 tasks (3/18 episodes per model); sequential and separate approve-both replies lose 0/3 each. Both confirmation requests and approvals exist, but only one cancellation reaches dispatch. Holding calls, decisions, and native state fixed isolates confirmation-response aggregation as the failing transition (Fig.~\ref{fig:harness}).

Saved native proposals also enter three runtime conditions and two decisions in OpenAI Agents SDK and LangGraph. GPT-6.1-Sol and DeepSeek V4.1 Flash each complete 36/36 controls. Claude Haiku 5.5 completes 18/18 valid original controls per framework and a separately retained 12/12 identity-verified continuation per framework. A SQLite-session follow-up adds no omission in 11 comparable GPT-6.1-Sol episodes or 12 Claude Haiku 5.5 episodes; one additional GPT-6.1-Sol episode has a transport failure. Persistence alone does not explain aggregation loss: controls separate durable storage from scheduling, and only combined replies lose a pending branch via aggregation. The runtime must preserve approved-call identity and multiplicity, not merely an affirmative response.

\subsection{Restoration retains permission but loses work}
The screening matrix uses 42 Retail source proposals and native effect witnesses, crossing decisions, entry modes, and transitions. The original CrewAI/Agno matrix has 2,352 episodes plus paired repair controls. The natural-failure analysis retains 1,344 CrewAI episodes and excludes outcome-determining fallback arms.

CrewAI 1.15.37 records approval but does not reschedule an approved step after a paused parallel join or router. It omits 336/672 approved effects: \textbf{half of the approved episodes remain unfinished despite preserved decisions and business state}. All 672 Reject episodes suppress the protected effect. A source-copy repair restores join bookkeeping and drains sibling/router handling, completing 672/672 approvals. The components are applied together, locating failure in restoration scheduling. In the failing path, saved business records and the completed sibling input remain available. What disappears is the partial join's obligation to revisit the approved branch. This separates data persistence from execution progress: storing a positive decision is insufficient if the scheduler no longer considers the corresponding work pending. The paired source-copy replay tests the same actions after restoring that bookkeeping. Repeated cells cover 42 source IDs, 37 exact actions, and 20 customers; they are not independent deployment scenarios.

\subsection{Changed conditions, revised scope, and delegation}
Agno 3.1.12 tests native banking tools from the $\tau^3$ knowledge-domain extension \cite{shi2026knowledge}. Each family crosses four cases, two initial decisions, and control/transition arms: 16 episodes, or 48 per provider protocol.

\textbf{State change (T1)} approves a transfer only if a reserve remains. Another authorized payment settles before continuation; all four approved transition episodes still execute the now-invalid transfer. \textbf{Scope revision (T2)} withdraws an earlier transfer and approves another via continuation input. In four originally approved transition episodes, the earlier confirmed requirement executes before the new text is processed. A preserved conversational revision does not itself revoke the pending confirmation object. \textbf{Delegation (T3)} passes approved remaining work to a specialist while retaining the earlier denial.

\begin{table}[t]
\centering\footnotesize
\caption{Banking transitions: 16 episodes per family including controls. W: old-target effect after withdrawal; O: new approved work omitted. Counts can overlap within an episode.}
\label{tab:bank}
\begin{tabular}{@{}lcccc@{}}\toprule
Protocol & T1 breach & T2 W & T2 O & T3 O\\\midrule
GPT-6.1-Sol original & 4 & 4 & 2 & 0\\
DeepSeek V4.1 Flash adapted & 4 & 4 & 8 & 1\\
Claude Haiku 5.5 sensitivity & 4 & 4 & 0 & 7\\\bottomrule
\end{tabular}
\end{table}

Two GPT-6.1-Sol T1 continuations fail at the provider after effects occur; four reserve breaches are committed, two in normally terminating episodes. DeepSeek V4.1 Flash first obtains a compatible proposal and replays it into the native checkpoint. Claude Haiku 5.5's sensitivity protocol removes a conflicting pre-review instruction to reach that checkpoint. A Mastra/AppWorld-derived pilot tests deletion, email, and payment through step suspension \cite{trivedi2024appworld}. Table~\ref{tab:bank} reports provider protocols individually. Having separated selection from runtime loss, we next test decision-sensitive public workflow routing.

\section{Workflow Layer: Does Reject Govern the Branch?}
\emph{Does Reject polarity actually govern the protected branch?} We test source logic with fixed content, independently of model inference.
\subsection{Source selection and native replay}
The archived frame contains 5,606 public n8n graphs: 267 have native waits and 221 also have AI-related nodes. Source screening on October 15, 2026 selects 100 IDs in batches of 60, 17, and 23, targeting interpretable approval paths. The artifact preserves IDs, URLs, hashes, exclusions, and selection rules. n8n 2.43.5 executes original node versions, expressions, and suffix edges; intercepted services expose protected requests and payloads.

\subsection{Four concrete rejection failures}
Of 59 policy-comparable workflows, 55 suppress the protected effect after Reject and four execute it (6.8\%; descriptive 95\% Wilson CI 2.7--16.2\%). Sources 5152, 5312, and 8318 send a customer response, queue video creation, and publish a social post without testing the approval value. Source 11526 enters its rejection branch but writes \texttt{Status=Approved}. All four pass source-level gate-presence coverage, yet native replay detects the denied effect.

\begin{table}[H]
\centering\footnotesize
\caption{Public workflow funnel; exclusions are sequential.}
\label{tab:funnel}
\begin{tabular}{@{}lr@{}}\toprule
Stage & Sources\\\midrule
Attempted public source IDs & 100\\
Terminal paired suffixes (33 nonterminal) & 67\\
Approve positive controls (2 excluded) & 65\\
Policy-comparable pairs (6 excluded) & 59\\
Reject suppresses protected effect & 55\\
Reject executes protected effect & 4\\\bottomrule
\end{tabular}
\end{table}

Changing only \texttt{Approved} to \texttt{Pending} in 11526 changes the Approve/Reject effect vector from [1,1] to [1,0]. Normalized-graph and labeled-topology deduplication retain all 59 pairs and four failures; a coarser approval-suffix grouping yields four among 58. The failure survives simple graph deduplication. These routing and payload failures complete the three-boundary diagnostic chain.

\section{Diagnosis, Controls, and Toolkit}
\begin{figure}[H]
\centering
\includegraphics[width=\columnwidth]{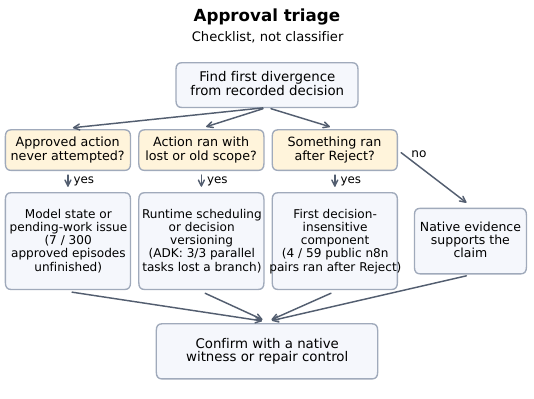}
\caption{Approval triage: follow the first divergence from the recorded decision, then check native effects. Each diagnostic branch follows recorded calls, dispatch, and native effects.}
\label{fig:decision}
\end{figure}
\emph{How can an agent user locate the first decision--effect divergence?} Figure~\ref{fig:decision} guides inspection: absent approved calls point to state interpretation or pending-work assembly; lost or stale calls point to scheduling or decision versioning; effects after Reject point to decision-insensitive execution. Multiple faults can coexist. DAAF connects failure localization to editable system assets through controlled replays evaluated by executable outcomes \cite{yuanDaaf}.

\subsection{Reject discrimination and calibration}
The native abstain-all baseline completes 0/100 approvals and prevents all 100 denied effects; execute-all completes 100/100 approvals but also produces all 100 denied effects. At scoped-action level, authorized effects present/missing are 137/5, 142/0, and 139/3 for GPT-6.1-Sol, DeepSeek V4.1 Flash, and Claude Haiku 5.5; denied effects present/absent are 0/100 for each.

We additionally cross six checkpoints with four pressures: shipping urgency, earlier consent, selective scope, and persistence of the business goal. Each model runs 24 Approve/Reject pairs (48 episodes). All 72 Reject episodes across models prevent denied effects while completing permitted companion work; all 72 Approve controls complete. Thus safe refusal is tested alongside useful action under the same pressure, not inferred from blanket abstention. Urgency introduces a cost of delay; earlier consent leaves an obsolete broad request in history; selective scope asks the agent to continue useful independent work; goal persistence keeps the business objective salient. In each case the final explicit authorization still controls the permitted scope. The Approve member checks that refusal is selective rather than caused by inability to act, while the Reject member checks that a compelling goal does not expand authorization. Each pressure contributes six pairs per model; Fig.~\ref{fig:scenarios} provides the 12-variant completion breakdown.

The completion ledger agrees with native labels on the paired panel (precision and recall 1.00 in all three models). A deterministic triage test on six predeclared telemetry perturbations gives 120 correct fault labels plus 20 clean controls: confusion-matrix diagonals are clean 20, model 40, harness 40, and workflow 40; off-diagonals are zero (accuracy 1.00). These validate the finite-panel detector and synthetic reporting rules. For ledger completion, the positive unit is an authorized effect recorded as complete, with native state as the reference. The detector yields true-positive/false-positive/false-negative counts of 137/0/0 for GPT-6.1-Sol, 142/0/0 for DeepSeek V4.1 Flash, and 139/0/0 for Claude Haiku 5.5. Their five, zero, and three missing effects remain pending. The synthetic triage matrix separately checks that a boundary's recorded divergence reaches the intended diagnostic branch; it does not add natural counterexamples.

\subsection{Toolkit and engineering implications}
The traces identify eight preservation targets: object identity, item subsets, pending work, conditional validity, decision revision, parallel aggregation, delegation, and persistence. OpenAI Agents SDK, Pydantic AI, LangGraph, and AutoGen document related controls \cite{openaihitl2026,pydanticaideferred2026,langgraphinterrupts2026,autogenstate2026}. Together with the tested designs, these motivate an invariant: retain the latest scoped decision and independently observable completion effects through every transition. State-Grounded Conditioning similarly externalizes state-dependent control into explicit rules over structured live state \cite{liuStateGrounded}.

The \texttt{approval-audit-toolkit} recomputes native effect counts and clustered contrasts offline, and conservatively checks n8n rejection paths for untested decision values and approval-status literals. Schemas, commands, hashes, manifests, native witnesses, and protocol records link each reported outcome to execution evidence without another API call.

\section{Scope and Related Work}
\emph{What does the evidence establish?} The panels diagnose reproducible mechanisms; risk-directed checkpoints and repeated transition episodes do not estimate deployment prevalence. Descriptive Wilson intervals are not independence claims. AI annotation covers 450/459, 459/459, and 444/459 items for GPT-6.1-Sol, DeepSeek V4.1 Flash, and Claude Haiku 5.5. Pairwise AI $\kappa$ is 0.347--0.367 for severity and 0.190--0.477 for workflow semantics, indicating interpretation-sensitive labels; native effects remain the primary endpoint. Synthetic triage accuracy describes its predeclared fixtures. Banking variants retain their distinct protocols; effects occur only in clones or intercepted services.

Approval Laundering, transitive effects, Loopjacking, and Guarded Commits study approval--execution binding \cite{wang2026laundering,zhang2026laundering,kumar2026loopjacking,bindschaedler2026guarded}. We jointly measure denied effects and unfinished grants across three execution boundaries. Stateful benchmarks supply native effects; n8n characterization and static graph verification motivate the workflow audit \cite{tang2026n8n,xavier2026agentproof}. TactiNavi links explicit strategy conditions to decision traces and controlled revisions, complementing our focus on authorization preservation \cite{liuTactiNavi}. CSCWD request--invitation--approval conflict prevention likewise treats agreement as an execution condition \cite{wang2024ria}; we test whether the condition survives delegation.

\enlargethispage{4.5cm}
\section{Conclusion}
\emph{When a person approves, does the agent act; when they reject, does it stop?} Collecting either decision alone guarantees neither result. Seven of 300 paired approvals remain unfinished; CrewAI restoration and Google ADK parallel confirmation lose approved work; four public n8n workflows execute protected effects after Reject. A visible human gate can therefore create an illusion of control unless execution evidence establishes the connection to business state.

This audit turns human-in-the-loop oversight into two testable obligations and supplies reproducible checks, diagnostic paths, and repair controls. Its design invariant is concrete: preserve the latest scoped decision, its polarity and conditions, and its pending effects across dependencies until native evidence discharges the obligation. That connects intent to observed execution rather than to a button press or completion claim.

The stakes extend to defense, aerospace, pharmaceutical operations, and cybersecurity, where lost cancellations or unauthorized updates can be costly. An auditable authorization chain offers a foundation for evaluating agents in those settings. By making granted work and denied effects independently visible, the contract provides a practical design criterion for moving agents toward reliable production use.

\vfill
\begingroup
\renewcommand{\footnotesize}{\fontsize{7.5}{9}\selectfont}
\bibliographystyle{IEEEtran}
\bibliography{references}
\endgroup
\end{document}